\documentclass{article}

\usepackage{microtype}
\usepackage{graphicx}
\usepackage{subfigure}
\usepackage{booktabs} 

\usepackage{hyperref}

\usepackage[accepted]{icml2024}

\usepackage{amsmath}
\usepackage{amssymb}
\usepackage{mathtools}
\usepackage{amsthm}

\usepackage[capitalize,noabbrev]{cleveref}

\theoremstyle{plain}

\theoremstyle{definition}

\theoremstyle{remark}

\usepackage{float}
\usepackage[retain-zero-uncertainty=true]{siunitx}
\usepackage[version=4]{mhchem}
\icmltitlerunning{Conditional Normalizing Flows for Active Learning of Coarse-Grained Molecular Representations}

\begin{document}

\twocolumn[
\icmltitle{Conditional Normalizing Flows for Active Learning\\of Coarse-Grained Molecular Representations}



\icmlsetsymbol{equal}{*}

\begin{icmlauthorlist}
\icmlauthor{Henrik Schopmans}{int,iti}
\icmlauthor{Pascal Friederich}{int,iti}
\end{icmlauthorlist}

\icmlaffiliation{int}{Institute of Nanotechnology, Karlsruhe Institute of Technology, Kaiserstr. 12, 76131 Karlsruhe, Germany}
\icmlaffiliation{iti}{Institute of Theoretical Informatics, Karlsruhe Institute of Technology, Kaiserstr. 12, 76131 Karlsruhe, Germany}

\icmlcorrespondingauthor{Pascal Friederich}{pascal.friederich@kit.edu}

\icmlkeywords{Machine Learning, ICML, coarse-graining, active learning, normalizing flow}

\vskip 0.3in
]



\printAffiliationsAndNotice{}  


\begin{abstract}
    Efficient sampling of the Boltzmann distribution of molecular
systems is a long-standing challenge. Recently, instead of generating long
molecular dynamics simulations, generative machine learning methods such as
normalizing flows have been used to learn the Boltzmann distribution directly,
without samples. However, this approach is susceptible to mode collapse and
thus often does not explore the full configurational space. In this work, we
address this challenge by separating the problem into two levels, the
fine-grained and coarse-grained degrees of freedom. A normalizing flow
conditioned on the coarse-grained space yields a probabilistic connection
between the two levels. To explore the configurational space, we employ
coarse-grained simulations with active learning which allows us to update the
flow and make all-atom potential energy evaluations only when necessary. Using
alanine dipeptide as an example, we show that our methods obtain a speedup to molecular 
dynamics simulations of approximately $15.9$ to $216.2$ compared to the speedup of $4.5$ of the 
current state-of-the-art machine learning approach.


\end{abstract}

\section{Introduction}\label{sec:introduction}
\begin{figure}[ht]
\vskip 0.2in
\begin{center}
\centerline{\includegraphics{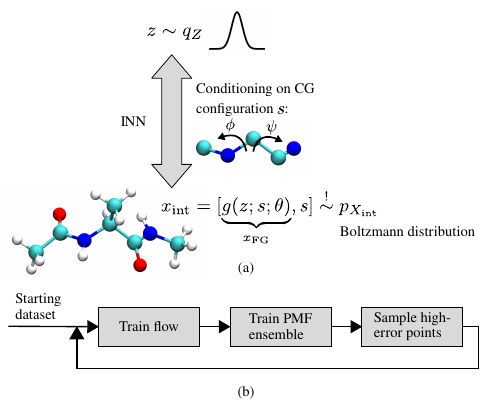}} \caption{(a) The
conditional normalizing flow transforms the latent variable $z$ of the latent
distribution to the target configuration $x_\text{int}$ conditioned on the CG
configuration $s$. (b) Illustration of the iterative three-step active learning
cycle.}
\label{fig:active_learning_overview}
\end{center}
\vskip -0.2in
\end{figure}

Coarse-graining (CG) of molecular dynamics (MD) simulations is a powerful
technique to bridge time and length scales that are often infeasible to study on
an atomistic level. Numerous processes, especially large-scale conformational
changes in biomolecules such as the folding of proteins, occur on the
microseconds to seconds timescale
\cite{spiritiDevelopmentApplicationEnhanced2012}. Since a typical time step used
in all-atom molecular dynamics simulations is
\sisetup{range-phrase=-,range-units=single}
\SIrange{1}{2}{\femto \second}, covering such time scales is currently 
computationally very expensive and infeasible for many systems of interest.

Bottom-up coarse-graining of all-atom simulations introduces effective sites
that describe the collective motion of one or multiple atoms of the all-atom
description. The potential of mean force (PMF) allows simulation of the CG
system and averages out many of the fast degrees of freedom, making the PMF
smoother than the all-atom potential energy
\cite{jinBottomupCoarseGrainingPrinciples2022}. This allows larger time steps to
be used than for typical all-atom simulations. The reduced number of degrees of
freedom, combined with a larger time step, allows for studying larger systems
with extended simulation times.

Next to classical CG force fields such as Martini
\cite{marrinkTwoDecadesMartini2023} or UNICORN
\cite{liwoScaleconsistentApproachDerivation2020}, machine learning (ML) models
have been successfully used to parametrize the PMF as a function of the CG
coordinates \cite{
husicCoarseGrainingMolecular2020,thalerDeepCoarsegrainedPotentials2022a,
zhangDeePCGConstructingCoarsegrained2018,
majewskiMachineLearningCoarsegrained2023, charronNavigatingProteinLandscapes2023, wangMachineLearningCoarseGrained2019}.
Typically, a long and thus very costly all-atom MD trajectory is required to learn the PMF on the CG
level. Although the CG model enables the simulation of significantly
longer simulation times, it is unlikely to reveal novel statistics not already
present in the all-atom training data, such as interesting transitions and new metastable states.

In this work, we demonstrate how active learning (AL) can be used to iteratively refine
the PMF while efficiently exploring the configurational space on the CG level.
This removes the need for long all-atom simulations to obtain the PMF.
Implementing an active learning workflow for coarse-grained simulations yields a
critical challenge: PMF values in high-error regions of the CG space cannot be
directly evaluated and need to be obtained from conditionally sampling the
all-atom Boltzmann distribution in the high-error CG configurations.
\citeauthor{duschatkoUncertaintyDrivenActive2024} \yrcite{duschatkoUncertaintyDrivenActive2024} showed that this can be
accomplished using constrained MD simulations. However, this approach is limited to
small systems since an MD simulation needs to be performed for each high-error
configuration. In this work, we use a simulation-free alternative based on a
conditional normalizing flow \cite{winklerLearningLikelihoodsConditional2023,xiaoMethodModelConditional2019,ardizzoneGuidedImageGeneration2019} to sample the all-atom level given a high-error CG configuration.

\citeauthor{noeBoltzmannGeneratorsSampling2019a}
\yrcite{noeBoltzmannGeneratorsSampling2019a} showed that a normalizing flow can
not only be trained with samples from MD, but also by using the Boltzmann
distribution $\sim\exp(-\frac{E(x)}{k_\mathrm{B} T})$ directly in the loss function without
samples (training by energy). This allows for simulation-free learning of the
potential energy surface. Recently, this idea has been extended in other
studies
\cite{mahmoudAccurateSamplingMacromolecular2022,midgleyFlowAnnealedImportance2023}.
However, this form of training a normalizing flow typically suffers from mode
collapse. While there are approaches that improve the tendency of mode collapse
when training without samples
\cite{midgleyFlowAnnealedImportance2023,felardosDesigningLossesDatafree2023,vaitlGradientsShouldStay2022},
a universal solution is yet to be found, especially for large systems or systems
with very large energy barriers.
The preprint by \citeauthor{zhangMachineLearnedInvertibleCoarse2023} \yrcite{zhangMachineLearnedInvertibleCoarse2023} used a
conditional normalizing flow for a coarse-graining task without active learning, however no benchmarks
or implementation details are available.

We solve the aforementioned challenges by developing a normalizing flow trained by
energy to generate the fine-grained degrees of freedom conditioned on the CG
space. We further show how, based on the trained conditional normalizing flow,
an ensemble of PMF models can be trained in the CG space. We iteratively sample
new high-error points from the PMF ensemble and subsequently refine the
conditional normalizing flow. Since the main modes of the Boltzmann distribution
are encapsulated in the CG space and the normalizing flow only learns the
conditional ``soft'' fine-grained degrees of freedom, we avoid mode collapse.

Using alanine dipeptide with a two-dimensional CG space consisting of the dihedral angles $\phi$ and $\psi$, we demonstrate that our methodology produces PMF maps of higher accuracy
while using approximately two orders of magnitude less potential energy evaluations compared to running all-atom MD simulations and one order of magnitude less compared to the state-of-the-art ML approach of learning the Boltzmann distribution without samples \cite{midgleyFlowAnnealedImportance2023}.
Our methods obtain a speedup to molecular dynamics
simulations of approximately $15.9$ to $216.2$ compared to the speedup of $4.5$ of the 
method by \citeauthor{midgleyFlowAnnealedImportance2023} \yrcite{midgleyFlowAnnealedImportance2023}.
To the best of our knowledge, this is the first time that active learning
on the CG side is demonstrated without the use of costly constrained MD
simulations to evaluate the PMF. We believe that extending this methodology to
more complicated CG spaces and molecular systems will yield a powerful technique
to efficiently generate coarse-grained potentials and simulations. Furthermore, while we focus on molecular systems in this work, our approach
can be applied to any general sampling problem of an unnormalized probability density, where
a meaningful CG space that contains the main modes of the probability distribution can be defined.

\section{Background}\label{sec:background}
\subsection{Normalizing Flows}
A normalizing flow transforms a latent probability distribution $q_Z(z)$, for
example, a standard Gaussian $\mathcal{N}\left(z ; 0, \mathbf{I} \right)$, using a transformation $x=g(z;\theta)$ with parameters $\theta$. The
transformed probability density $ q_X(x;\theta) $ can be expressed using the change of
variables formula \cite{dinhNICENonlinearIndependent2015}:

\begin{align}
    &q_X(x;\theta)=q_Z\left(f(x;\theta)\right)\left|\operatorname{det} J_{x \mapsto z}\right| \label{eq:norm_flow_transform} \\
    &\text{with the Jacobian } J_{x \mapsto z} = \frac{\partial f(x;\theta)}{\partial x^T}
\end{align}

This requires the inverse of $ g(z;\theta) $, $ f(x;\theta) \equiv z =
g^{-1}(x;\theta) $. A popular choice to parametrize such an invertible function
(invertible neural network, INN) is a stack of coupling layers, where the
dimensions of the input $ x_{1:D} $ are split into two parts, $ x_{1:d} $ and
$ x_{d+1: D}$. The first part undergoes an invertible transformation conditioned on the second part,
which stays unchanged (see SI Section \ref{SI:sec:coupling_layers} for details).

The parameters of the flow can be trained in such a way that the generated
distribution $q_X(x;\theta)$ approximates the Boltzmann distribution $p_X(x) $.
Normalizing flows have a vital advantage over most other generative machine
learning methods in that they can express the exact probability density efficiently. This allows training both with
samples from the target distribution using the forward Kullback–Leibler divergence (KLD) and without samples using the reverse
KLD.\footnote{Recently, diffusion models have also been used for training with the reverse KLD \cite{jingTorsionalDiffusionMolecular2022}.}

To derive both modes of training, we first define the following probability distributions:
\begin{itemize}
    \item $q_Z(z)$ is the latent distribution we sample from.
    \item $q_X(x;\theta)$ is the generated distribution of the flow (see Equation \ref{eq:norm_flow_transform}).
    \item $p_X(x) \sim \exp \left( -E(x) / (k_{\mathrm{B}} T) \right)$ is the target
    Boltzmann distribution, where $ E(x) $ is the potential energy of the molecular
    system, with temperature $T$ and Boltzmann constant $k_{\mathrm{B}}$. 
    \item $p_Z(z;\theta) = p_X(g(z;\theta)) \left|\operatorname{det} J_{z \mapsto x}\right|$ is the Boltzmann probability transformed by the INN to the
    latent space.
\end{itemize}

\paragraph{Training by Example.}
Using the forward KLD as the loss function, we can train a normalizing flow
using samples from the target distribution
\cite{noeBoltzmannGeneratorsSampling2019a}:

\begin{align}
    &\mathrm{KL}_{\theta}\left[p_X \| q_X\right] =C-\int p_X(x) \log q_X(x ; \theta) \, \mathrm{d} x \\
    & =C-\mathbb{E}_{x \sim p_X}{\left[\log q_Z\left(f(x;\theta)\right) + \log \left|\operatorname{det} J_{x \mapsto z}\right| \right]}
    \label{eq:flow_training_by_example}
\end{align}

\paragraph{Training by Energy.}


We can also use the reverse KLD to train the flow directly using the target
energy function of the system without samples from the target density \cite{noeBoltzmannGeneratorsSampling2019a}:

\begin{align}
    &\mathrm{KL}_{\theta}\left[q_Z \| p_Z\right] =C-\int q_Z(z) \log p_Z(z ; \theta) \, \mathrm{d} z \\
    & =C-\mathbb{E}_{z \sim q_Z} \left[\underbrace{\log p_X\left(g(z;\theta)\right)}_{-\frac{1}{k_\mathrm{B} T} E(g(z;\theta))+C_1} + \log \left|\operatorname{det} J_{z \mapsto x}\right| \right]
    \label{eq:flow_training_by_energy}
\end{align}

In this way, one samples from the flow and adjusts the weights of the
transformation to fit the Boltzmann distribution given by the energy function.

\subsection{Coarse-Graining}

Coarse-graining combines the atom coordinates $ x \in \mathbb{R}^{3N} $ of a
molecular system into $m<3N$ coordinates using a CG mapping $ s=\xi(x)\text{, }
s \in \mathbb{R}^m $. Typically, the coordinates $s$ are obtained from a linear
transformation $ \mathbb{R}^{3N} \rightarrow \mathbb{R}^{m = 3M} $. A popular
choice is a ``slicing'' mapping \cite{yangSlicingDicingOptimal2023} that simply
uses a selection $M$ of the $N$ all-atom coordinates as CG coordinates, such as the
coordinates of the backbone atoms in a protein. Here, we will consider general,
potentially nonlinear CG mappings.\footnote{Low-dimensional nonlinear CG mappings are
often called collective variables in the literature.}

Our goal is to construct a potential of mean force (PMF) $U_\text{PMF}(s)$ that
can be used to sample from the CG space, e.g., through Langevin dynamics or
Metropolis Monte Carlo (MC). To ensure thermodynamic consistency in the CG
configurational space we need
\cite{wangMachineLearningCoarseGrained2019,noidMultiscaleCoarsegrainingMethod2008}

\begin{equation}
U_\text{PMF}(s) \equiv-k_\mathrm{B} T \ln \underbrace{ \frac{\int p_X(x) \delta(s-\xi(x)) \, \mathrm{d} x}{\int p_X(x) \, \mathrm{d} x} }_{p^{\mathrm{CG}}(s)} \, \text{.}
\label{eq:pmf_def}
\end{equation}

Due to the form of Equation \ref{eq:pmf_def}, the potential of mean force is also called the 
(conditional) free energy.

Unfortunately, we cannot directly train a model to predict the potential of mean
force, since training labels are not directly available. We can, however,
use the all-atom forces projected to the CG space, $h(x)$. The expectation value of these
projected forces for a given CG configuration yields the negative
gradient of the PMF (CG mean force), which can be derived directly from Equation \ref{eq:pmf_def} \cite{kalligiannakiGeometryGeneralizedForce2015, noidMultiscaleCoarsegrainingMethod2008,izvekovMultiscaleCoarseGrainingMethod2005}:

\begin{align}
-\nabla_s U_\text{PMF}(s) = \mathbb{E}_{x \sim p_X}[h(x) \mid \xi(x)=s]
\label{eq:mean_force}
\end{align}

If one uses a ``slicing'' CG mapping, the projected all-atom forces $h(x)$ are simply the
all-atom forces of the chosen CG atoms. One can, however, also find an expression for
$h(x)$ in other scenarios, even in the general nonlinear case
\cite{kalligiannakiGeometryGeneralizedForce2015}.

Evaluating the conditional expectation value in Equation \ref{eq:mean_force} is
not straightforward. Instead, one can show that minimizing the surrogate loss
\cite{noidMultiscaleCoarsegrainingMethod2008, izvekovMultiscaleCoarseGrainingMethod2005}

\begin{equation}
    \chi(W)=\left\langle\|h(x)+\nabla_s U_\text{PMF}(\xi(x) ; W)\|^2\right\rangle_{x \sim p_X}
    \label{eq:force_matching}
\end{equation}

also yields the correct $U_\text{PMF}$ model with parameters $W$. This is called multiscale
force-matching.

\citeauthor{kohlerFlowMatchingEfficientCoarseGraining2023}
\yrcite{kohlerFlowMatchingEfficientCoarseGraining2023} recently showed an
alternative  to the force-matching method called ``flow-matching'' that uses a
normalizing flow trained directly on the distribution of all-atom MD
samples transformed to the CG space. The PMF of the CG representation is then given by $U_\text{PMF}(s)
= -k_{\mathrm{B}} T \ln p^\text{CG}(s)$, where $p^\text{CG}(s)$ can be obtained directly from
the normalizing flow using Equation \ref{eq:norm_flow_transform}.

Both approaches, multiscale force-matching and flow-matching, require a
comprehensive all-atom dataset from the Boltzmann distribution for the training of the PMF.

\section{Methods}\label{sec:methods}
Now, we show how an iterative active learning workflow for coarse-graining of
molecular systems can be constructed without the need to perform long and
costly all-atom simulations.

We define an internal coordinate representation
$x_\text{int}(x)=[x_\text{FG}(x),\xi(x)]$ (with $[\,]$ denoting concatenation), which splits
the all-atom cartesian coordinates $x$ into fine-grained coordinates $x_\text{FG}(x)$ and coarse-grained
coordinates $s=\xi(x)$. $ x_\text{FG}(x) $ are the remaining fine-grained coordinates needed
to reconstruct $ x $ given $ s $, $x=x([x_\text{FG},s])$.\footnote{If the potential
energy is rotation- and translation-invariant, the orientation and translation
do not have to be reconstructed.} In the case of ``slicing'' mappings, this can
be a simple splitting of the atom coordinates, but it can also be an
actual internal coordinate representation, which we use for our experiments.

A normalizing flow can now be used to parametrize the probability distribution
of the fine-grained coordinates conditioned on the CG coordinates (see also
Figure~\ref{fig:active_learning_overview}):

\begin{align}
q_{X_{\mathrm{FG}}}\left(x_{\mathrm{FG}} \mid s ; \theta\right)
=q_Z(f(x_\text{FG} ; s; \theta))\left|\operatorname{det} J_{x_\text{FG} \mapsto z}\right|
\end{align}

Training by energy of this conditional flow can be performed using the loss in Equation~\ref{eq:flow_training_by_energy} in the
space $X=X_\text{FG}$ with conditioning on $s$. Here, the target distribution is given by 

\begin{align}
& p_{X_\text{FG}}(x_\text{FG} \mid s) \sim \\ 
& \frac{1}{p^\text{CG}(s)} \underbrace{\exp \left(
\frac{-E(x([g(z;s;\theta),s]))}{k_\mathrm{B} T} \right) \left|\operatorname{det} J_{x_\text{int} \rightarrow x} \right| }_{\sim p_{X_\text{int}}(x_\text{int})} \, \text{,} \notag
\end{align}

where $p^\text{CG}(s)$ is independent of flow parameters and can be absorbed into the constant of the reverse KLD loss.


\subsection{Active Learning Workflow}
We start with an initial dataset obtained from a short all-atom simulation. This
is used to initially train the conditional normalizing flow by example (Equation \ref{eq:flow_training_by_example}).
The starting dataset will typically only cover a small fraction of the full configurational space of the system.\footnote{One can,
of course, also use two starting datasets of distinct minima to find transition paths in the CG space.}
Furthermore, the CG configurations $ s=\xi(x) $ of the starting dataset
are used as the initial dataset of the active learning workflow. Then, each
active learning iteration consists of the following steps (see also Figure~\ref{fig:active_learning_overview}b):

\begin{enumerate}
\item We first train the conditional normalizing flow by energy (Equation \ref{eq:flow_training_by_energy}). 
For the conditioning, we use the $N$ high-error CG
configurations added at the end of the previous AL iteration (or from the starting dataset in
the first iteration). Additionally, in each epoch, we select $\gamma \cdot
N$ with $\gamma=0.3$ random CG configurations from older iterations. This stabilizes the flow in areas of
previous iterations and gives flexibility to still adapt slightly if needed.

\item We then train an ensemble of $10$ PMF models. The training strategy to
obtain the PMF models is described in Section~\ref{sec:obtaining_PMF}.

\item In the last step of each iteration, we sample points in the CG space that
exceed a defined threshold of the ensemble standard deviation using Metropolis Monte
Carlo (MC). We uniformly broaden the obtained high-error points by sampling
uniformly in a hypersphere around them in CG space. The broadened points are added to
the AL dataset, where \SI{80}{\percent} are used for training, and
\SI{20}{\percent} as test samples.

\end{enumerate}

After finishing each AL iteration, we stop the active learning workflow if the
forward KLD of the PMF using the test dataset is smaller than a defined threshold or the trajectories
of the MC sampling reach a specified maximum length.
The normalizing flow is initialized only once at the beginning of the AL workflow 
and is then progressively updated during the AL iterations.
The PMF ensemble is reinitialized every iteration.

Training a normalizing flow by energy to match the Boltzmann distribution of a
molecular system can be difficult, since high-energy configurations, e.g., due
to clashes between atoms, can yield very high loss values (Equation
\ref{eq:flow_training_by_energy}). This is especially problematic in the
beginning of training, but also later, since the flow will never exactly
resemble the Boltzmann distribution without any clashes. Previous works
\cite{noeBoltzmannGeneratorsSampling2019a,midgleyFlowAnnealedImportance2023}
mitigated the problem by using a regularized potential energy function that
applies a logarithm above a threshold $E_\text{high}$ and cuts off the energy above a threshold
$E_\text{max}$. We found empirically that in our case removing a few of the highest loss values from each
batch of samples yields more stable experiments than energy regularization.
Whether this procedure also improves the training of non-conditional flows may be explored
in future work.

\subsection{Grid Conditioning}
If the chosen CG space is low-dimensional, it is also
possible to simply uniformly cover the CG space instead of running the proposed
AL workflow (see discussion in Section~\ref{sec:discussion}). Therefore, we further show the results of training a conditional normalizing flow by energy on a grid in the CG space. In these experiments, we again first train
by example on the starting dataset and then by energy using the CG configurations
of the grid as conditioning.

\subsection{Obtaining the Potential of Mean Force}
\label{sec:obtaining_PMF}

Obtaining the PMF for coarse-grained simulations is typically done using
force-matching or flow-matching as described in Section~\ref{sec:background}.
Since the samples from the trained flow approximately follow the Boltzmann distribution, 
we could use force-matching to train our PMF models. In
our scenario, however, since we have access to the probability distribution
learned by the conditional normalizing flow $ q_{X_{\mathrm{FG}}}\left(x_{\mathrm{FG}} \mid s ; \theta\right) $, we
can alternatively obtain the PMF values directly using its definition in
Equation~\ref{eq:pmf_def}. We will discuss the advantages of this approach in
Section~\ref{sec:discussion}. As we show in SI Section~\ref{SI:sec:obtaining_F}, we can express the PMF using the
following expectation value:

\begin{align}
    U_\text{PMF}(s) = -k_\mathrm{B} T \ln \mathbb{E}_z\left[ \frac{p_{X_\text{int}}\left([g(z;s;\theta),s]\right)}{q_{X_{\mathrm{FG}}}\left(g(z;s;\theta) \mid s ; \theta\right)} \right] \, ,\label{eq:F_main} \\
    p_{X_\text{int}}(x_\text{int}) \sim \exp \left(
    -\frac{E(x([g(z;s;\theta),s]))}{k_\mathrm{B} T} \right) \left|\operatorname{det} J_{x_\text{int} \rightarrow
    x} \right|
\end{align}

This allows us to get the PMF $U_\text{PMF}(s)$ for a given CG configuration $s$ by
sampling from the conditional normalizing flow and evaluating the potential
$E(x)$ to obtain $p_{X_\text{int}}(x_\text{int})$. In order to obtain accurate 
PMF values, we only need sufficient overlap between the distribution of the conditional
normalizing flow and the Boltzmann distribution, since Equation \ref{eq:F_main} includes
a form of reweighting.

While training the flow by energy, we hold $k=30$ copies of each CG configuration in
the AL dataset. Each time one of the copies is selected in a batch when training
the flow by energy, we store the evaluated
potential energy and the probability $q_{X_{\mathrm{FG}}}\left(x_{\mathrm{FG}} \mid s ; \theta\right)$ 
for that copy. This means that we
always store the latest $k=30$ samples for each CG configuration. This allows us 
to train an ensemble of PMF models at the end of each active learning iteration without
making additional potential energy evaluations. Subsequently, this PMF ensemble can then
be used to sample new high-error configurations for the next iteration.

We want to emphasize that it is also possible to train a PMF model without
evaluating the expectation value in Equation~\ref{eq:F_main} explicitly. This
approach uses a surrogate loss function to match the free energies and can be
derived using a generalization of the multiscale force-matching proof. It is discussed in
more detail in SI Section~\ref{SI:sec:surrogate}. In SI Section~\ref{SI:sec:alternative_F} 
we further present an alternative to Equation~\ref{eq:F_main} that has been previously used by
\citeauthor{zhangMachineLearnedInvertibleCoarse2023} \yrcite{zhangMachineLearnedInvertibleCoarse2023}. 
While we obtained the most accurate results with our approach, a systematic comparison of the two
approaches is needed in future work.

\section{Experiments}\label{sec:experiments}
\subsection{Müller-Brown Potential}
\label{sec:experiments:mb}

\begin{figure}[h] 
\vskip 0.2in
\begin{center}
\centerline{\includegraphics[width=\columnwidth]{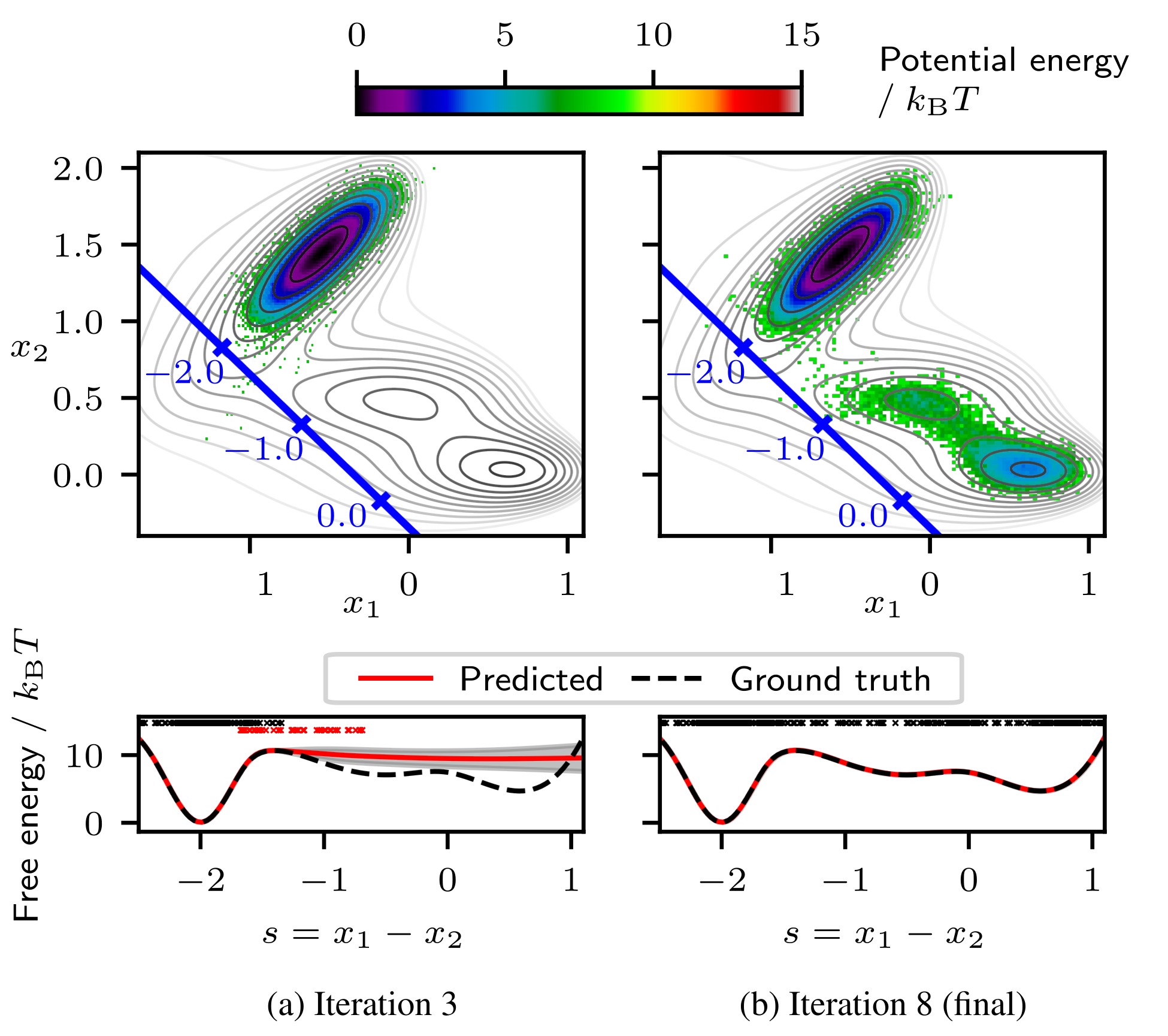}}
\caption{Visualization of two exemplary iterations of the AL workflow applied to
the Müller-Brown system. Bottom: PMF and its standard deviation. Training points 
from previous AL iterations are marked as black ``x'' at the top of the PMF, and new high-error points added in the current AL iteration are marked as red ``x''.
Top: Backmapped potential energy $\ln q_{S_\perp}(s_\perp \mid s; \theta) p^\text{CG}(s)$ from cascaded sampling
of the PMF and sampling of the flow. The blue axis represents the 1D CG coordinate
$s$. The fine-grained coordinate $s_\perp$ is orthogonal to the CG coordinate. See Figure~\ref{fig:SI:mb_steps}
in SI for a visualization of all iterations of this experiment.}
\label{fig:mb_al_steps}
\end{center}
\vskip -0.2in
\end{figure}

We first test the AL workflow on the 2D Müller-Brown potential
\cite{mullerLocationSaddlePoints1979} (Figure~\ref{fig:mb_al_steps} (top, contour lines)), starting 
the exploration in the global minimum. As a CG
mapping, we use the \SI{45}{\degree} rotated coordinate axis $s=\xi(x)=x_1-x_2$ (blue axis
in Figure~\ref{fig:mb_al_steps} (top)). The normalizing flow describes the
conditional probability distribution $q_{S_\perp}(s_\perp \mid s; \theta)$, where $s_\perp=x_1+x_2$ is
the ``fine-grained'' coordinate orthogonal to $s$. This relatively simple setup
serves as a first proof-of-concept of our methodology.

Since most coupling layers cannot directly be used to transform 1D probability
densities, we do not use a conventional normalizing flow, but simply transform
the latent distribution $ z \sim \mathcal{N}(0, 1) $ using the following
conditional transformation:
\begin{align}
s_\perp = z \cdot \text{NN}_\text{scale}(s) + \text{NN}_\text{mean}(s)
\end{align}
Here, $ \text{NN}_\text{scale} $ and $ \text{NN}_\text{mean} $ are fully
connected neural networks. 
This simple transformation suffices
to obtain an accurate potential of mean force since our approach of obtaining it 
(Section~\ref{sec:obtaining_PMF}) merely requires overlap of the distribution of the flow 
with the target distribution.

Figure \ref{fig:mb_al_steps} shows two exemplary iterations of an AL experiment
in the Müller-Brown system. One can see that after the final iteration 8, the
learned PMF (bottom, red) is almost identical to the ground-truth PMF (bottom, black).

Our AL experiments required \num{1.13(04)e5} potential energy
evaluations with \num{4.01(0.95)e5} MC steps in the CG potential
and obtained a forward KLD of the PMF of \num{2.04(23)e-4} (16 experiments
were performed). We further compare the AL workflow with the conventional CG
approach, where one first samples from the potential (here, using MC) and
subsequently extracts a PMF. Since the CG space of the Müller-Brown system is
only 1D, this is done using a histogram of the CG configurations in the MC
simulation\footnote{In higher-dimensional CG spaces one typically uses
force-matching or flow-matching to obtain the PMF, as discussed in Section~\ref{sec:background}.}.
We performed an MC simulation with \num{1e6} steps which yields 
a significantly larger KLD of the PMF of \num{1.02e-2} compared to our AL experiments. This comes from poor convergence of the MC simulation
since this simulation length only yields zero to a few transitions (more details in SI).
The AL workflow
learns the PMF more accurately using 10 times less potential energy evaluations
compared to the ``all-atom'' MC simulation of \num{1e6} steps.

\subsection{Alanine Dipeptide}
\label{sec:experiments:aldp}

Now, we consider a more complex scenario where we explore the configurational
space of the $22$-atom molecule alanine dipeptide using the
CG space $s=(\phi, \psi)^\top$ (see Figure~\ref{fig:active_learning_overview} for a visualization of the molecule and
definition of the dihedral angles $\phi$ and $\psi$). We use the same fully
internal coordinate representation as \citeauthor{midgleyFlowAnnealedImportance2023} \yrcite{midgleyFlowAnnealedImportance2023},
where the molecule is described using $ 3\cdot22-6=60 $ internal coordinates
(bond distances, angles, and dihedral angles). This representation obeys the
symmetries of the potential energy function since it is invariant to
translations and rotations of the molecule.

The conditional normalizing flow parametrizes the probability distribution
$q_{x_\text{FG}}(x_\text{FG} \mid s; \theta)$, where $x_\text{FG}$ are the remaining $58$ internal
fine-grained degrees of freedom. Similarly to
\citeauthor{midgleyFlowAnnealedImportance2023} \yrcite{midgleyFlowAnnealedImportance2023} we use a normalizing flow built from 12
monotonic rational-quadratic spline coupling layers
\cite{durkanNeuralSplineFlows2019} with residual networks as parameter networks
(details in SI Section~\ref{SI:sec:aldp:architecture}). Dihedral angles of freely rotating
bonds are treated as circular coordinates
\cite{rezendeNormalizingFlowsTori2020a,midgleyFlowAnnealedImportance2023}.
Furthermore, the circular coordinates in the input of the parameter network and
the periodic conditioning variables $\phi$ and $\psi$ of the parameter network
are treated using the periodic representation $ (\cos \eta, \sin \eta)^\top $
for each periodic variable $\eta$. During training, we further filter the
chirality of each batch and only train on structures in the L-form
\cite{midgleyFlowAnnealedImportance2023}.

\begin{figure*}[htb]
\vskip 0.2in
\begin{center}
\centerline{\includegraphics[width=\textwidth]{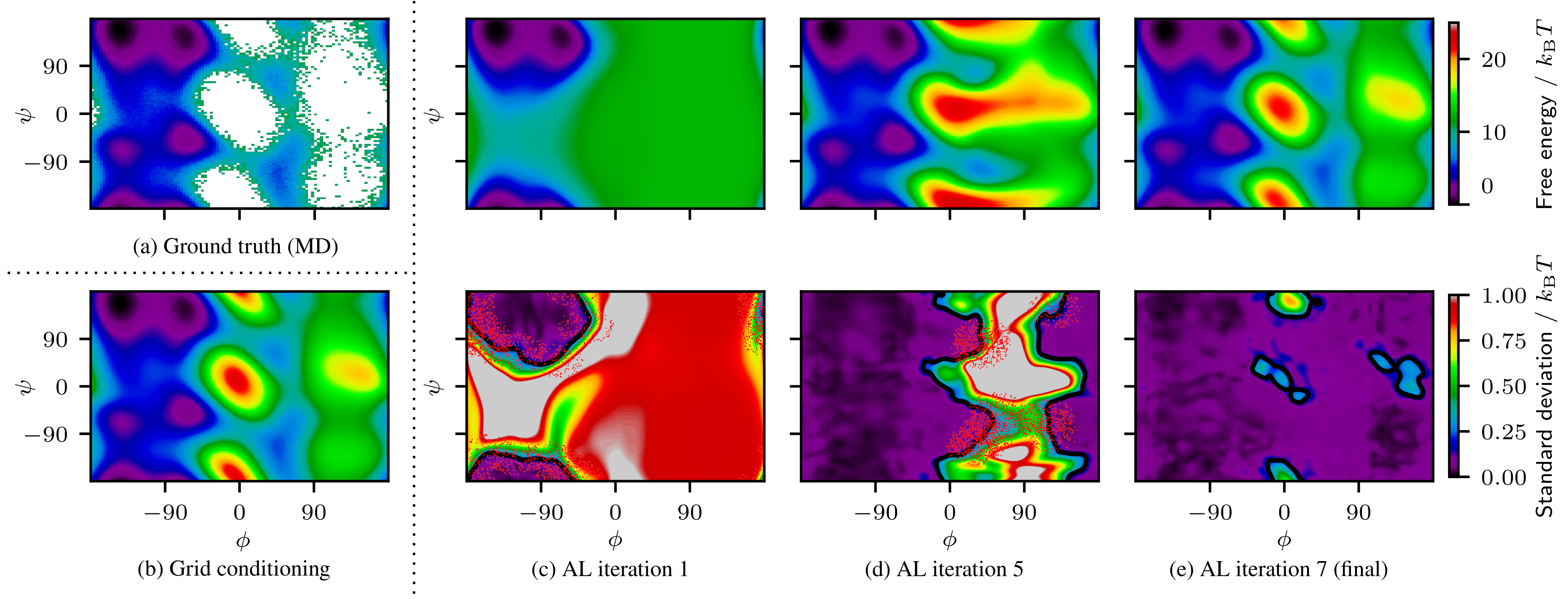}}
\caption{(a) Ground truth PMF of dihedral angles $\phi$ and $\psi$ of alanine
dipeptide from MD test dataset with \num{2.3e10} steps. (b) PMF from
grid conditioning experiment after \num{2.4e7} steps. (c-e) Three exemplary AL steps for the alanine
dipeptide system. Top: PMF of $\phi$ and $\psi$ at the end of the AL iteration.
Bottom: Standard deviation of the PMF. The contour line of the threshold in the
standard deviation when sampling new points ($0.2 \, k_\mathrm{B} T$) is drawn using a black
line. Newly added points after sampling using the PMF of this iteration are
shown as red dots.}
\label{fig:aldp_AL_steps}
\end{center}
\vskip -0.2in
\end{figure*}

Figure~\ref{fig:aldp_AL_steps}c-e shows the PMF for different iterations of an
active learning experiment applied to alanine dipeptide. Figure
\ref{fig:aldp_AL_steps}b shows the PMF obtained using a grid conditioning
experiment on a $100$x$100$ grid in the CG space. As one can see visually, the
resulting PMF after the last iteration of AL and the PMF from the grid
conditioning experiment are not only identical to the ground truth PMF (Figure
\ref{fig:aldp_AL_steps}a), but in addition cover regions that were not sampled at all in the reference MD simulation. 
Especially in transition regions of high energy, e.g.,
at $\phi \approx \SI{120}{\degree}$, our obtained PMF is much more detailed
and complete than the PMF from the ground truth MD dataset, where even after \num{2.3e10} steps, large parts of the high-energy regions are completely missing. 


We now consider a quantitative comparison of our approach with previous methods. Table~\ref{table:aldp_results} includes the results of
\citeauthor{midgleyFlowAnnealedImportance2023} \yrcite{midgleyFlowAnnealedImportance2023}, where annealed importance sampling
with $\alpha$-divergence with $\alpha=2$ (Flow AIS Bootstrap) was used to directly train a
normalizing flow by energy to sample from the configurational space of alanine
dipeptide. This is currently the only publication that succeeds in learning the full
Boltzmann distribution of alanine dipeptide with a generative model without samples and without mode collapse. 
Furthermore, we include the results of training a normalizing flow 
with the reverse KLD \cite{midgleyFlowAnnealedImportance2023}, where mode collapse can be 
observed. Besides these ML approaches, we also include a comparison
with a separate MD simulation for two MD simulation lengths. As the main metric to judge the accuracy of the PMF, we use the forward KLD of the PMF calculated on the test dataset.

In addition to the forward KLD of the PMF, we also provide metrics for the generated all-atom distribution: The all-atom log-likelihood
$\mathbb{E}_{p_X(x)} \log q_X(x ; \theta)$ and the reverse effective sample size (ESS) (see Section \ref{SI:sec:metrics} for details). We find the all-atom log-likelihood is similar across all methods (except for the flow with reverse KLD training, where mode collapse is observed), while Flow AIS Bootstrap \cite{midgleyFlowAnnealedImportance2023} achieves better reverse ESS compared to our results. However, the main goal of this study is to obtain an accurate PMF model that can be used to run coarse-grained simulations. Therefore, we now focus on the forward KLD of the PMF.

\begin{figure}[h]
\vskip 0.2in
\begin{center}
\centerline{\includegraphics{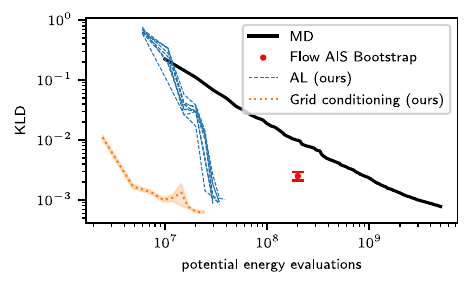}}
\caption{Forward KLD of the PMF of alanine dipeptide as a function of the number
of potential energy evaluations. 
For grid conditioning and Flow AIS Bootstrap we additionally show the standard error.}
\label{fig:KL_divergence_over_steps}
\end{center}
\vskip -0.2in
\end{figure}

Next to Table~\ref{table:aldp_results}, the forward KLD of the PMF as a function of the number of potential energy evaluations is further visualized in Figure~\ref{fig:KL_divergence_over_steps}. We find that the forward KLD of the PMF as a function of the number of potential energy evaluations of our methods decreases substantially faster than Flow AIS Bootstrap and MD.
Compared to the Flow AIS Bootstrap \cite{midgleyFlowAnnealedImportance2023}
results, our AL workflow uses approximately one order of magnitude less potential
energy evaluations while obtaining a substantially smaller KLD. The experiments with grid
conditioning show similar results while reaching the final Flow AIS Bootstrap KLD almost two orders of magnitude faster. When running MD simulations, the KLD is significantly higher, even
when using $2-3$ orders of magnitude more potential energy evaluations (\num{1e9})
(see also Figure \ref{fig:KL_divergence_over_steps}). As discussed before, our method yields (accurate) PMF maps also in regions where no ground truth data is available. This makes the quantification of the increase in accuracy difficult. Therefore, we quantify the speedup of our method at a fixed accuracy (KLD value): A KLD of \num{2.5e-3} (Flow AIS Bootstrap) is reached after approximately \num{9.1e8} MD steps (reference), \num{2e8} steps in Flow AIS Bootstrap (speedup of $4.5$), \num{2.9e7} potential energy evaluations + \num{2.8e7} CG MC steps (speedup of $15.9$), and \num{4.2e6} grid conditioning steps (speedup of $216.2$). Here, for the AL workflow, we generously counted the cost of one CG step to be equal to one all-atom potential energy evaluation. We discuss the sampling efficiency in the CG space in more detail in Section~\ref{sec:discussion}.
We assume that the accuracy reached by our methods is higher than that of the available ground truth data, making the aforementioned numbers lower bounds of the actual speedup values. 

We note that at the accuracies we are achieving here, the reported KLD is not anymore a representative measure of the increasing accuracy of the PMF, because (a) it is logarithmic so the accuracy of the PMF in high-energy regions is influencing the KLD in a negative exponential way, and (b) we do not have any good ground truth data as reference in the KLD, specifically in those high energy regions. That means we cannot quantitatively measure if our PMF becomes more accurate than the PMF shown in Figure~\ref{fig:aldp_AL_steps}a, while visual inspection of Figure~\ref{fig:aldp_AL_steps}e and ~\ref{fig:aldp_AL_steps}b clearly indicates that we achieve substantially improved PMF maps compared to the MD-derived ground truth, at a fraction of the cost.


\begin{table*}[htb!]
\caption{Results of the experiments with alanine dipeptide. We compare
the performance of different approaches: A normalizing flow trained with the reverse KLD \cite{midgleyFlowAnnealedImportance2023}, Flow AIS Bootstrap with buffer \cite{midgleyFlowAnnealedImportance2023}, MD simulations of two different
lengths, our active learning workflow, and our grid conditioning experiments - once fully converged and once with fewer number of evaluations.
As the main metric to estimate the accuracy of the PMF we use the forward KLD of the PMF calculated on the test dataset. Additionally, we provide metrics for the generated all-atom distribution: The all-atom log-likelihood 
$\mathbb{E}_{p_X(x)} \log q_X(x ; \theta)$ and the reverse effective sample size (ESS).
For the given metrics, $\uparrow$ indicates higher is better, $\downarrow$ indicates lower is better. For our experiments, we provide the mean and standard error over 8
experiments. Note that the AL experiments required, additionally to the all-atom potential energy
evaluations, \num{3.01(0.83)e7} MC steps.}
\label{table:aldp_results}
\vskip 0.15in
\begin{center}
\begin{small}
\begin{sc}

\begin{tabular}{lllll}
\toprule
Method & Pot. energy & Forward & \multicolumn{2}{c}{all-atom metrics $\uparrow$} \\
& evaluations $\downarrow$ & KLD PMF $\downarrow$ & $\mathbb{E}_{p_X(x)} \log q_X(x ; \theta)$ & Reverse ESS / \% \\

\midrule

Flow with reverse KLD & \num{2.5e8} & \num{3.15(19)} & \num{100(32)} & \num{54(12)} \\
Flow AIS Bootstrap & \num{2e8} & \num{2.51(39)e-3} & \textbf{\num{211.54(000)}} & \textbf{\num{92.8(1)}} \\
MD (long) & \num{1e9} & \num{2.32e-3} & - & - \\
MD (short) & \num{1e8} & \num{1.87e-2} & - & - \\
AL (ours) & \num{3.35(10)e7} & \num{9.29(24)e-4} & \num{211.18(04)} & \num{43.83(3.81)} \\
Grid cond. (ours, converged) & \num{2.40e7} & \textbf{\num{6.32(28)e-4}} & \num{211.49(01)} & \num{83.92(1.15)} \\
Grid cond. (ours, short) & \textbf{\num{4.85e6}} & \num{1.68(13)e-3} & \num{210.93(02)} & \num{25.58(1.02)} \\

\bottomrule
\end{tabular}

\end{sc}
\end{small}
\end{center}
\vskip -0.1in
\end{table*}

\section{Discussion}\label{sec:discussion}
Methods such as all-atom MD or normalizing flows trained by energy struggle with
correctly describing rare transition regions of molecular systems, since they
are not sampled frequently in the Boltzmann distribution. Our approach of using
a normalizing flow conditioned on CG coordinates circumvents this problem due to
the conditional sampling of the Boltzmann distribution. This allows the correct
description of the PMF even in very high-energy regions. Furthermore, using the
proposed AL workflow, the configurational space can be explored more efficiently
in the lower-dimensional smoother PMF of the CG space compared to all-atom simulations.

We need to emphasize that in the low-dimensional CG spaces of the two systems
that our work covers, one can simply sample the CG coordinates on a grid, as
shown in Table~\ref{table:aldp_results} for alanine dipeptide.
In higher-dimensional CG spaces, other means of sampling the CG space instead of
using a grid become strictly necessary, e.g., using Langevin dynamics or Metropolis Monte Carlo-based exploration. 
As discussed in detail in SI Section~\ref{SI:sec:aldp_performance}, already for our example alanine dipeptide - where implicit solvation energy and force evaluations are very fast - the surrogate PMF model can be sampled faster and thus more efficiently than the all-atom level.

In the future, we envision our approach to be used for more complex systems and CG mappings.
This includes conventional CG mappings, such as using only the backbone atoms of a molecule as CG 
coordinates, where Langevin dynamics can be used to sample in the CG space and find
new high-error configurations. While the transition barriers of alanine dipeptide are
relatively low and it is possible to sample sufficiently with all-atom MD,
applying the active learning methodology to systems with large energy barriers
will allow efficient exploration of the configurational space in a relatively
low-dimensional space. When using density functional theory or an active-learned ML surrogate thereof on the all-atom level, sampling the lower-dimensional PMF will be much faster than sampling the all-atom level.

Scaling to higher-dimensional problems, such as using C$_\alpha$ CG mappings of proteins, will yield new challenges. Since the reverse KLD is mode-seeking, it can only be used to learn distributions that are approximately unimodal. It needs to be seen if the conditional side-chain distribution of proteins and other systems of interest fulfills this constraint. Potentially, other optimization objectives that are not mode-seeking, such as the $\alpha$-divergence with $\alpha=2$ used by \citeauthor{midgleyFlowAnnealedImportance2023} \yrcite{midgleyFlowAnnealedImportance2023}, need to be used instead of the reverse KLD.

Also non-conventional, nonlinear CG mappings such as folding coordinates for proteins or
known reaction coordinates in chemical reactions might be interesting for
efficient sampling of the configurational space using our proposed methodology -
either with active learning or using the grid conditioning strategy. This can
yield accurate PMF profiles along collective variables of interest. \citeauthor{zhangMachineLearnedInvertibleCoarse2023} \yrcite{zhangMachineLearnedInvertibleCoarse2023} already showed first results in this direction, though performance comparisons
and implementation details are missing. If the
all-atom coordinates cannot be directly reconstructed in such a way that it
obeys the conditioning (which is the case for most nontrivial nonlinear reaction coordinates), one needs
to introduce an additional consistency loss term to make sure that $\xi(x)$ of the generated
configurations is equal to the conditioning $s$ (see
\cite{zhangMachineLearnedInvertibleCoarse2023}).

To scale the shown methodology to larger systems in the future, some recent advances in the architecture 
of normalizing flows can be used \cite{midgleySEEquivariantAugmented2023, kohlerEquivariantFlowsExact2020, garciasatorrasEquivariantNormalizingFlows2021, draxlerFreeformFlowsMake2023},
which includes continuous normalizing flows with equivariant graph neural networks. 
Moving away from the here-used internal coordinate representation further allows the usage of one 
conditional normalizing flow across multiple molecular systems. This would allow the transfer of a 
pre-trained normalizing flow and CG potential to novel systems, where the active learning workflow 
can be used to further refine the CG potential when necessary.

Besides the application in the active learning workflow, the approach of learning the conditional
probability $p_{X_\text{FG}}(x_\text{FG} \mid s)$ and then extracting the PMF (see Section~\ref{sec:obtaining_PMF})
might have advantages in data efficiency over both multiscale force-matching and flow-matching even in the 
non-active-learning scenario (which requires a comprehensive all-atom
MD dataset). Further details can be found
in the SI in Section~\ref{SI:sec:comparison_methods}.

\section{Conclusion}\label{sec:conclusion}
In this work, we showed how conditional normalizing flows can be used to build an
active learning workflow for coarse-grained simulations. 
When testing our method with the 22-atom molecule alanine dipeptide, we demonstrated a speedup to molecular dynamics 
simulations of approximately 15.9 to 216.2 compared to the speedup of 4.5 of the
current state-of-the-art ML approach by \citeauthor{midgleyFlowAnnealedImportance2023} \yrcite{midgleyFlowAnnealedImportance2023}.
We obtain a higher accuracy using approximately an order of magnitude less potential energy
evaluations compared to \citeauthor{midgleyFlowAnnealedImportance2023} \yrcite{midgleyFlowAnnealedImportance2023}. 
Compared to performing all-atom MD simulations to extract the PMF we achieve higher accuracy
with two orders of magnitude less potential energy evaluations.
Since all current ML coarse-graining approaches require such an extensive MD simulation to train the PMF, our active learning approach outperforms them in terms of data efficiency.

We see our work as a first demonstration of the potential that lies in performing active
learning in the coarse-grained space of molecular systems. For the first time,
this is possible without first needing a long all-atom trajectory or constrained
MD simulations. Our results show that if one can construct a (potentially nonlinear)
lower-dimensional CG space that includes the main modes of a molecular system, a conditional
normalizing flow can efficiently learn this conditional probability distribution
- either with uniform coverage of the CG space if it is low-dimensional, or with
an explorative active learning approach for higher-dimensional CG spaces. We
are confident that this will boost the utility and application of
machine-learned coarse-grained potentials and hope to see further developments
in this direction.

\section*{Data and Code Availability}\label{sec:data}
Our reference implementation of the described active learning workflow can be found on \url{https://github.com/aimat-lab/coarse-graining-AL} (v1.0). Code to reproduce all experiments is provided.

\section*{Acknowledgements}\label{sec:acknowledgements}
The authors would like to thank the anonymous reviewers for
their valuable comments and suggestions. H.S. acknowledges financial support by the German Research Foundation (DFG) through the Research Training Group 2450 “Tailored Scale-Bridging Approaches to Computational Nanoscience”.
P.F. acknowledges funding from the Klaus Tschira Stiftung gGmbH and the pilot program Core-Informatics of the Helmholtz Association (HGF).
The authors acknowledge support by the state of Baden-Württemberg through bwHPC. Parts of this work were performed on the HoreKa supercomputer funded by the Ministry of Science, Research and the Arts Baden-Württemberg and by the Federal Ministry of Education and Research.

\section*{Impact Statement}\label{sec:impact}
This paper presents work whose goal is to advance the field of Machine Learning. There are many potential societal consequences of our work, none of which we feel must be specifically highlighted here.

\bibliography{bib}
\bibliographystyle{icml2024}

\newpage
\appendix
\onecolumn
\section{Appendix}

\subsection{Application to New Problems}
\label{SI:sec:new_problems}

In this manuscript, we focused on the application of our coarse-graining approach to the sampling
of molecular tasks. However, the described workflow (both the active learning and the grid conditioning approach)
can be applied to any sampling problem of an unnormalized probability density, where a meaningful CG space
that includes the main modes of the distribution can be defined. For example, our method could also be applied
to sampling problems of lattice field theories, such as the $\phi^4$ theory \cite{nicoliEstimationThermodynamicObservables2021}.

The main steps to apply our workflow to a new system are the following:
\begin{itemize}
    \item Implement the target log probability (energy) function and the CG mapping function (see \emph{System} class in our source code)
    \item Decide upon a normalizing flow architecture, such as RNVP or rational-quadratic splines
    \item Implement a PMF model architecture that incorporates potential symmetries of the CG space
    \item Generate a small starting dataset using Metropolis Monte Carlo or other sampling techniques
    \item If needed, update the active learning sampler that is used to obtain new high-error configurations. For example,
    when applying our method to higher-dimensional molecular systems with backbone CG mappings, one might
    consider Langevin dynamics to sample the CG space.
    \item Update config options to match the system details
\end{itemize}

\subsection{Obtaining the Potential of Mean Force}
\label{SI:sec:obtaining_F}

Here, we derive Equation \ref{eq:F_main} of the main text which we use
to train the PMF models.

We start with the definition of the potential of mean force:

\begin{align}
    U_\text{PMF}(s) &= -k_\mathrm{B} T \ln \left[ \int_x \exp \left( - \beta E(x) \right) \delta \left( s - \xi(x) \right) \, \mathrm{d} x \right] + C_0 \\
    &\underset{C_0=0}{=} -k_\mathrm{B} T \ln \left[ \int_x Z \cdot p_X(x) \delta \left( s - \xi(x) \right) \, \mathrm{d} x \right] \\
    &= -k_\mathrm{B} T \ln \left[ \int_x \frac{q_{X_{\mathrm{FG}}}\left(x_{\mathrm{FG}} \mid s ; \theta\right)}{q_{X_{\mathrm{FG}}}\left(x_{\mathrm{FG}} \mid s ; \theta\right)} p_X(x) \delta \left( s - \xi(x) \right) \, \mathrm{d} x \right] + C_1 \\
    &\underset{C_1=0}{=} -k_\mathrm{B} T \ln \left[ \int_z q_Z(z) \frac{1}{q_{x_{\mathrm{FG}}}\left(g(z;s;\theta) \mid s ; \theta\right)} p_{X_\text{int}}([g(z;s;\theta),s]) \, \mathrm{d} z \right] \\
    &= -k_\mathrm{B} T \ln \mathbb{E}_{z \sim q_Z} \underbrace{\left[ \frac{p_{X_\text{int}}\left([g(z;s;\theta),s] \right)}{q_{x_{\mathrm{FG}}}\left(g(z;s;\theta) \mid s ; \theta\right)} \right]}_{\equiv G_1(s,z)}
\end{align}

Here, $p_{X_\text{int}}(x_\text{int}) \sim \exp \left(
    -\frac{E(x(x_\text{int}))}{k_\mathrm{B} T} \right) \left|\operatorname{det} J_{x_\text{int} \rightarrow
    x} \right|$ .

If the distribution of the conditional normalizing flow sufficiently overlaps with the target Boltzmann distribution, this expectation value can be used to obtain accurate PMF values for training of the PMF ensemble.

\subsubsection{Alternative Formulation}
\label{SI:sec:alternative_F}

Here, we describe an alternative to Equation \ref{eq:F_main} to obtain the PMF from the conditional normalizing flow. Again, we
start with the definition of the PMF:

\begin{align}
    U_\text{PMF}(s) &= - k_\mathrm{B} T \ln \left[ \underbrace{\int_x \exp \left( - \beta E(x) \right) \delta \left( s - \xi(x) \right) \, \mathrm{d} x}_{\equiv Z(s)} \right] + C \underset{C=0}{=} - k_\mathrm{B} T \ln Z(s) \\
    &= \int_x \underbrace{\delta \left( s - \xi(x) \right) \frac{\exp(-\beta E(x))}{Z(s)}}_{p(x \mid s)} E(x) \, \mathrm{d} x + k_\mathrm{B} T \int_x \underbrace{\delta \left( s - \xi(x) \right) \frac{\exp(-\beta E(x))}{Z(s)}}_{p(x \mid s)} \ln \left[ \underbrace{\frac{\exp(-\beta E(x))}{Z(s)}}_{p(x \mid s)} \right] \, \mathrm{d} x \\
    &= \int_x p(x \mid s) E(x) \, \mathrm{d} x + k_\mathrm{B} T \int_x p(x \mid s) \ln p(x \mid s) \, \mathrm{d} x \\
    &= \int_{x_\text{int}} p(x_\text{int} \mid s) E(x(x_\text{int})) \, \mathrm{d} x_\text{int} + k_\mathrm{B} T \int_{x_\text{int}} p(x_\text{int} \mid s) \ln p(x(x_\text{int}) \mid s) \, \mathrm{d} x_\text{int} \\
    &\approx \int_z q_Z(z) E(x(\underbrace{[g(z;s;\theta),s]}_{x_\text{int}})) \, \mathrm{d} z + k_\mathrm{B} T \int_z q_Z(z) \ln \left[ q_{X_\text{FG}} (g(z;s;\theta) \mid s; \theta) \cdot \left|\operatorname{det} J_{x \rightarrow x_\text{int}} \right| \right] \, \mathrm{d} z \\
    &= \mathbb{E}_{z \sim q_Z} \underbrace{\left[ E(x(\underbrace{[g(z;s;\theta),s]}_{x_\text{int}})) + k_\mathrm{B} T \ln \left( q_{X_\text{FG}} (g(z;s;\theta) \mid s; \theta) \cdot \left|\operatorname{det} J_{x \rightarrow x_\text{int}} \right| \right) \right]}_{\equiv G_2(s,z)} \label{eq_sup:F_expectation_alternative}
\end{align}

This formula is the same as used by \citeauthor{zhangMachineLearnedInvertibleCoarse2023}
\yrcite{zhangMachineLearnedInvertibleCoarse2023}. In the same way as Equation
\ref{eq:F_main}, it allows us to estimate the PMF $U_\text{PMF}(s)$ of a CG configuration
$s$ using the conditional normalizing flow
$q_{X_{\mathrm{FG}}}\left(x_{\mathrm{FG}} \mid s; \theta\right)$. In
practice, we found that this alternative formula does not work as well as using
Equation \ref{eq:F_main} directly. However, further comparisons
of the different approaches need to be done in the future.

\subsubsection{Surrogate Loss}
\label{SI:sec:surrogate}

Here, we show that it is also possible to obtain the PMF without
explicitly evaluating the expectation value in Equation \ref{eq:F_main} or
Equation \ref{eq_sup:F_expectation_alternative}. When using a mean squared error
loss, we can show that we will obtain the correct PMF even without the
contraction along $z$. This is analogous to how we can estimate the expectation value of
the projected all-atom forces in multiscale force-matching using a mean squared error
surrogate loss. This approach can be applied to both the expectation value in
Equation \ref{eq:F_main} and the alternative version in Equation
\ref{eq_sup:F_expectation_alternative}.

In general, we want to obtain the following expectation value:

\begin{align}
   H_i(s) = \mathbb{E}_{z \sim q_Z} G_i(s,z)
\end{align}

Then, $U_\text{PMF}(s)=-k_\mathrm{B} T \ln H_1(s)$ for version 1 and $U_\text{PMF}(s)=H_2(s)$ for the alternative version 2.

We can formulate a surrogate loss to train a model $H(s; W)$ with
parameters $W$ to match the expectation value:

\begin{align}
    \chi(W)    &= \langle \left[ G_i(s,z) - H(s; W) \right]^2 \rangle_{s,z} \\
            &= \langle \left[ G_i(s,z) - \langle G_i(s,z) \rangle_z + \langle G_i(s,z) \rangle_z - H(s; W) \right]^2 \rangle_{s,z} \\
            &= \underbrace{\langle \left[ G_i(s,z) - \langle G_i(s,z) \rangle_z \right]^2 \rangle_{s,z}}_{\equiv \chi_\text{Noise}}
            + \underbrace{\langle \left[ \langle G_i(s,z) \rangle_z - H(s; W) \right]^2 \rangle_{s,z}}_{\equiv \chi_\text{Expectation}}\\
            &+ \underbrace{2 \langle (G_i(s,z) - \langle G_i(s,z) \rangle_z)(\langle G_i(s,z) \rangle_z - H(s; W)) \rangle_{s,z}}_{\equiv \chi_\text{Mixed}}
\end{align}

As one can see, the surrogate loss $\chi$ can be decomposed into three parts:
First, there is a noise term $\chi_\text{Noise}$, which is independent of the free
energy model parameters $W$. This is equivalent to the noise term that we
find in the multiscale force-matching objective \cite{wangMachineLearningCoarseGrained2019}. Furthermore, there is our main objective
$\chi_\text{Expectation}$. This is the loss term that we actually need, as it
matches $H(s; W)$ onto the expectation value $\langle G_i(s,z) \rangle_z$.

We now have to show that the mixed term $ \chi_\text{mixed} $ is zero, which is
possible because the distribution of $z$ is independent of $s$:

\begin{align}
    \frac{\chi_\text{mixed}}{2}&=
        \langle \langle G_i(s,z) \langle G_i(s,z) \rangle_z \rangle_z \rangle_s
        - \langle \langle \langle G_i(s,z) \rangle_z \langle G_i(s,z) \rangle_z \rangle_z \rangle_s \\
        &- \langle \langle G_i(s,z) H(s; W) \rangle_z \rangle_s + \langle \langle \langle G_i(s,z) \rangle_z H(s; W) \rangle_z \rangle_s \\
        & = 0
\end{align}

As one can see, we can obtain the PMF even when not explicitly contracting along $z$. In
practice, we found that contraction should still be done if possible, as it
lowers the noise significantly and makes training much easier. However, in
scenarios in which one does not want to evaluate the same CG configuration in
the flow multiple times, one can still use the mean squared error to obtain the
correct PMF.

\subsection{Invertible Neural Networks: Coupling Layers}
\label{SI:sec:coupling_layers}

As discussed in the main text, a normalizing flow requires an invertible function approximator
(invertible neural network, INN), which can be constructed using a stack of coupling layers.
Here, the dimensions of the input $ x_{1:D} $ are split into two parts, $ x_{1:d} $ and
$ x_{d+1: D}$ (usually, this splitting is done using a random mask). 
The first part is transformed elementwise conditioned on the
second part, while the second part is unchanged
\cite{midgleySEEquivariantAugmented2023}:

\begin{align}
x^\prime_{1: d} & = B\left(x_{1: d} ; x_{d+1: D}\right), \\
x^\prime_{d+1: D} & =x_{d+1: D} .
\end{align}

If $ B $ is invertible (monotonic) and since $x_{d+1: D}$ is unchanged
and thus given when transforming forward and backward, this yields an invertible
architecture. The Jacobian matrix of such a coupling transform is lower
triangular and thus the Jacobian determinant can be efficiently computed using the
diagonal elements of the Jacobian matrix \cite{durkanNeuralSplineFlows2019}. 

\subsection{Comparison with Multiscale Force-Matching and Flow-Matching}
\label{SI:sec:comparison_methods}

Besides the active learning workflow, we want to shortly discuss a possible
application of the here-described approach to train PMF models in the
non-active-learning scenario, where training data from MD trajectories is
present. Here, our approach of learning $p_{X_\text{FG}}(x_\text{FG} \mid s)$ and then extracting the
PMF (see Section \ref{sec:obtaining_PMF}) has potential advantages over both multiscale force-matching and flow-matching.

Multiscale force-matching requires large amounts of training data since transition regions
need to be sufficiently covered to get a good gradient estimate. If this is not
the case, distinct minima on the PMF surface appear biased in energy. Flow
matching mitigates this problem by not working with solely gradient information. $p^\text{CG}(s)$ is
directly (or indirectly in the case of student-teacher training
\cite{kohlerFlowMatchingEfficientCoarseGraining2023}) learned from the
distribution of the training data in the CG space. However, this requires a
well-converged MD trajectory where different minima are correctly occupied.
Otherwise, the flow will learn this wrong occupation. 

In our here-described approach, both the shortcomings of multiscale force-matching and flow-matching are
addressed. By learning $p_{X_\text{FG}}(x_\text{FG} \mid s)$ from the training data, one does not depend
on gradient information alone. Furthermore, since $p^\text{CG}(s)$ is not directly
obtained from the occupation of the different minima in the dataset, the occupations do not
have to be fully converged in the trajectory. We only need enough training data
in the regions of interest (where the PMF should be estimated) to learn $p_{X_\text{FG}}(x_\text{FG}
\mid s)$. In this way, one can use our approach in the non-active-learning
scenario, where training data in the form of an MD trajectory is present. Closer
investigation and comparison of the data efficiencies and accuracies between the different
approaches is not the focus of this work and should be investigated in future
work.

\subsection{Müller-Brown Potential}
\label{SI:sec:MB}

\subsubsection{Potential Energy}
\label{SI:sec:MB:potential_def}

\begin{figure}[h]
\vskip 0.2in
\begin{center}
\centerline{\includegraphics{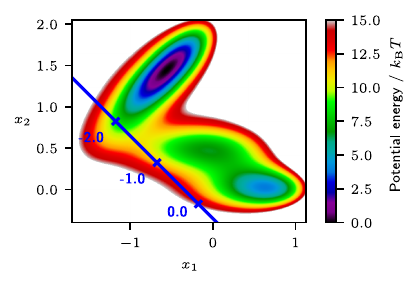}}
\caption{Müller-Brown potential. The blue axis represents the 1D CG coordinate
$s$. The fine-grained coordinate $s_\perp$ is orthogonal to the CG coordinate.}
\label{fig:SI:mb_potential}
\end{center}
\vskip -0.2in
\end{figure}

The Müller-Brown potential is defined through the following formula
\cite{mullerLocationSaddlePoints1979} (see Figure \ref{fig:SI:mb_potential} for a visualization):

\begin{align}
    & E_\text{MB}(x_1, x_2)=\sum_{i=1}^4 A_i \exp \left[a_i\left(x_1-\bar{x}_i\right)^2+\right. \\
    &\left.b_i\left(x_1-\bar{x}_i\right)\left(x_2-\bar{y}_i\right)+c_i\left(x_2-\bar{y}_i\right)^2\right] \notag \\
    \text{with} \quad & A=(-200,-100,-170,15) ; \quad a=(-1,-1,-6.5,0.7) \\
    & b=(0,0,11,0.6) ; \quad c=(-10,-10,-6.5,0.7) \\
    & \bar{x}=(1,0,-0.5,-1) ; \quad \bar{y}=(0,0.5,1.5,1) .
\end{align}

We use $\beta=\frac{1}{k_\mathrm{B} T}=0.1$ to evaluate the PMF and run the MC sampling of
the AL workflow.

Since the Müller-Brown potential is only 2D, the ground-truth PMF of our CG coordinate $s_\perp=x_1-x_2$
can be simply obtained using numerical integration of Equation \ref{eq:pmf_def}. Using this ground-truth PMF, 
the forward KLD values reported in Section \ref{sec:experiments:mb} have been calculated in the range 
$s \in [-2.5, 1.1]$ using a grid of 100 points.

\subsubsection{Starting Dataset}
To form the initial starting dataset for the Müller-Brown system we use a short MC
simulation starting in the lower minimum of the potential ($x_1=-0.25$, $x_2=1.5$). We
perform 500 MC steps with a Gaussian proposal distribution of scale $0.2$.
We use 100 randomly chosen unique positions from this MC dataset as the 
starting dataset of the AL workflow.

\subsubsection{Architecture}
\label{SI:sec:MB:architecture}

\paragraph{Normalizing Flow.}

As described in the main text, we use the following 1D transformation with the
latent distribution $ z \sim \mathcal{N}(0, 1) $ for the Müller-Brown
system:

\begin{align}
s_\perp = z \cdot \text{NN}_\text{scale}(s) + \text{NN}_\text{mean}(s)
\end{align}

The subnets $\text{NN}_\text{scale}$ and $\text{NN}_\text{mean}$ are fully
connected neural networks with layer dimensions $[ 1,64,64,1 ]$ and sigmoid
hidden layer activation functions. We use a standard scaler $s^\prime =
\frac{s-\bar{s}}{\sigma_s}$ based on the starting dataset of the AL workflow for the input of the two neural networks.

\paragraph{PMF.}
To predict the PMF and its standard deviation, we use an ensemble of 10 fully
connected neural networks with layer dimensions $[1, 64, 64, 1]$ and sigmoid hidden
layer activation functions. We use a standard scaler $s^\prime =
\frac{s-\bar{s}}{\sigma_s}$ calculated from the current dataset of the respective AL iteration.

To improve ensemble diversity, we initialize the weights and biases of each model in the ensemble using a Gaussian 
of mean 0 and standard deviation $\nu$. $\nu$ is sampled uniformly in the range $[0.1,3.0]$ for each model in the ensemble.

\subsubsection{Hyperparameters}

\paragraph{Normalizing Flow.}
We use the Adam optimizer \cite{kingmaAdamMethodStochastic2017} to train the normalizing flow.

When training by example on the starting dataset, we use a batch size of $16$ and
a learning rate of \num{5e-4}. 

When training the flow by energy, we use a batch size of 8 and a learning rate
of \num{5e-3}. We further clip gradients above a gradient norm of $20$. The first AL
iteration trains by energy for $12$ epochs, all subsequent iterations use $7$ epochs.

\paragraph{PMF.}
To train the PMF ensemble, we use the Adam optimizer with a learning rate of $0.001$
and a batch size of $5$. Training is performed for $1000$ epochs.

We use the bagging strategy to select the training data of each model in the ensemble.

\paragraph{Monte Carlo Sampling.}
We perform Monte Carlo Sampling in the CG space to find new high-error
configurations. We use a Gaussian proposal distribution of scale $0.1$ and an
error threshold of $0.4 \, k_\mathrm{B} T$.

In each AL iteration, we search for $1$ high-error point using $50$ MC
trajectories in parallel. The trajectories have a minimum length of $10$ steps, before
which we do not accept high-error configurations. The trajectories of the first AL iteration start in
the global minimum of the potential. All trajectories of subsequent iterations
start at the high-error configuration of the previous iteration. Each high-error
point sampled using MC is subsequently broadened with a uniform distribution of
width $1.0$.

In this way, we sample $65$ points for each high-error configuration, resulting
in $1\cdot65=65$ added CG configurations in each AL iteration. Since each CG configuration
has $30$ copies in the dataset (see Section \ref{sec:obtaining_PMF}), this yields \num{1950} new points.
\SI{80}{\percent} are subsequently used 
for training by energy in the next iteration, and \SI{20}{\percent} for calculation of a test loss.

If any of the trajectories reaches a length of \num{30000} steps, we stop the active 
learning workflow. The reported total number of MC steps do not include the steps of this final fixed-length MC
exploration that terminates the workflow due to reaching the maximum specified length.

\begin{figure*}[ht]
\vskip 0.2in
\begin{center}
\centerline{\includegraphics[width=1.0\textwidth]{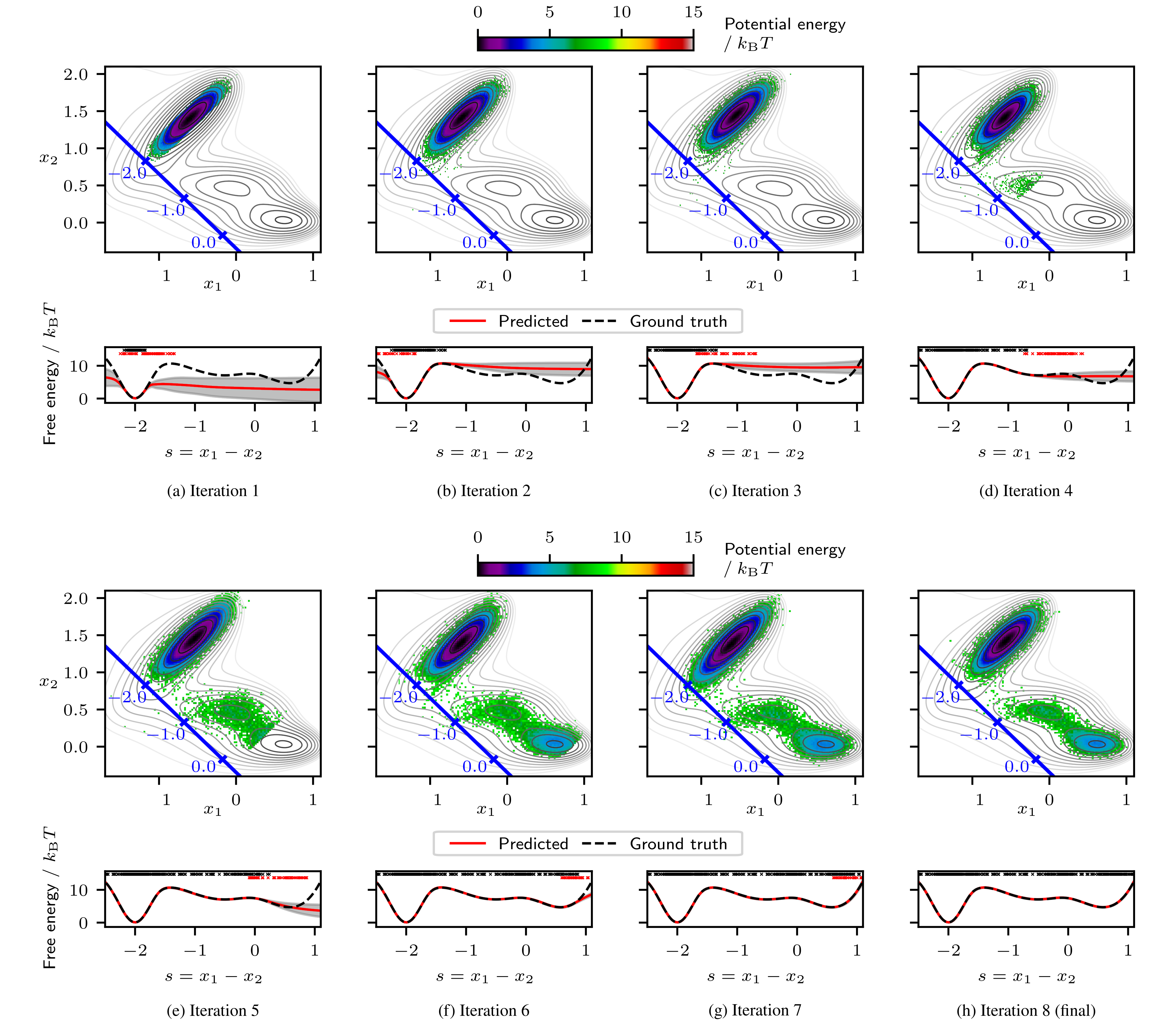}}
\caption{Visualization of all iterations of the AL workflow applied to
the Müller-Brown system. Bottom: PMF and its standard deviation. Training points 
from previous AL iterations are marked as black ``x'' at the top of the PMF, and new high-error points added in the 
current AL iteration are marked as red ``x''.
Top: Backmapped potential energy $\ln q_{S_\perp}(s_\perp \mid s; \theta) p^\text{CG}(s)$ from cascaded sampling
of the PMF and sampling of the flow.}
\label{fig:SI:mb_steps}
\end{center}
\vskip -0.2in
\end{figure*}

\begin{figure*}[ht]
\vskip 0.2in
\begin{center}
\centerline{\includegraphics[width=0.6\textwidth]{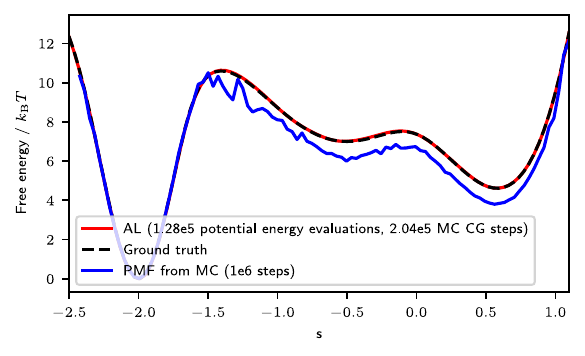}}
\caption{PMF as obtained using the AL experiment from Figure
\ref{fig:SI:mb_steps} (red), by numerical integration (ground truth, black), and from
an ``all-atom'' MC simulation with \num{1e6} steps (blue). The PMF from the ``all-atom'' MC simulation
is visibly biased, while the AL workflow yields an almost perfect PMF estimate.}
\label{fig:SI:mb_AL_MC_comparison}
\end{center}
\vskip -0.2in
\end{figure*}

\subsection{Alanine Dipeptide}
\label{SI:sec:aldp}

\subsubsection{Force-Field and Ground Truth Simulations}
\label{SI:sec:aldp:datasets}

We used the force-field AMBER ff96 with OBC GB/SA for implicit solvation
\cite{d.a.caseAmber20232023} for the ground truth simulations and potential energy
evaluations during the training of the normalizing flow. Energy evaluations and
simulations were performed using OpenMM 8.0.0 with the reference platform
\cite{eastmanOpenMMRapidDevelopment2017}. All simulations were performed at the
temperature \SI{300}{\kelvin} and included an iterative energy minimization (gradient descent) before starting the 
simulations from the obtained minimum energy configuration.

As a ground-truth test dataset, we used the dataset provided by
\citeauthor{stimperAlanineDipeptideImplicit2022} \yrcite{stimperAlanineDipeptideImplicit2022}
and \citeauthor{midgleyFlowAnnealedImportance2023} \yrcite{midgleyFlowAnnealedImportance2023},
which was generated using replica exchange MD simulations with a total of \num{2.3e10} potential energy and force evaluations. 

Additionally to this test dataset, we created an additional MD trajectory of length
\SI{5}{\micro \second} with time step \SI{1}{\femto \second} (\num{5e9} potential energy and force evaluations) to create the MD
entries in Table \ref{table:aldp_results} and the visualization in Figure
\ref{fig:KL_divergence_over_steps}.

\subsubsection{Evaluation Metrics} \label{SI:sec:metrics}
\paragraph{Forward KLD.} The ground truth test dataset mentioned in the previous section was used
to calculate the forward KLD of the PMF on a 100x100 grid (see KLD values in Table \ref{table:aldp_results}).

\paragraph{All-Atom Log-Likelihood.} The all-atom log-likelihoods in Table \ref{table:aldp_results} have been calculated using the ground truth test dataset as
\begin{align}\mathbb{E}_{p_X(x)} \log q_X(x ; \theta)=\mathbb{E}_{p_X(x)} \log \left(q_{X_{\mathrm{FG}}}\left(x_{\mathrm{FG}} \mid s ; \theta\right) \cdot p(s) \cdot\left|\operatorname{det} J_{x \rightarrow x_{\mathrm{int}}}\right|\right) \, \text{.}
\end{align}
Here, $p(s) \sim \exp \left( -\frac{1}{k_\text{B} T} U_\text{PMF}(s) \right) $ was obtained from the PMF and normalized on a 500x500 grid in the CG space.

\paragraph{ESS.} The reverse effective sample size (ESS) in Table \ref{table:aldp_results} has been calculated in the following way \cite{midgleySEEquivariantAugmented2023, martinoEffectiveSampleSize2017}:
\begin{align}
    &n_{\mathrm{e}, \mathrm{rv}}=\frac{1}{\sum_{i=1}^n \bar{w}\left(x_i\right)^2} \quad \text { with } \quad x_i \sim q_X\left(x_i; \theta\right) \, \text{,} \\ 
    & \bar{w}(x_i) = \frac{w(x_i)}{\sum_{i=1}^n w(x_i)} \text{ and } w(x_i) = \frac{p_X(x_i)}{q_X(x_i; \theta)}
\end{align}
Analogous to the all-atom log-likelihood calculation, we used $q_X(x ; \theta)=q_{X_{\mathrm{FG}}}\left(x_{\mathrm{FG}} \mid s ; \theta\right) \cdot p(s) \cdot\left|\operatorname{det} J_{x \rightarrow x_{\mathrm{int}}}\right|$, where $p(s) \sim \exp \left( -\frac{1}{k_\text{B} T} U_\text{PMF}(s) \right)$ was again obtained from the PMF and normalized on a 500x500 grid in the CG space. To sample from $q_X(x;\theta)$, we employed rejection sampling in the CG space and subsequently used the conditional normalizing flow to obtain all-atom samples. We used \num{1e7} samples drawn from $q_X(x;\theta)$ to estimate the reverse ESS. Analogous to \citeauthor{midgleyFlowAnnealedImportance2023} \yrcite{midgleyFlowAnnealedImportance2023}, due to outliers in the importance weights, we clipped the \num{1e3} highest importance weights to the lowest value among them.

\subsubsection{Coordinate Transformation}
We use a fully internal coordinate representation with $ 3\cdot22-6=60 $
internal coordinates (bond distances, angles, and dihedral angles). The internal
coordinates are shifted using the minimum energy reference
structure obtained using gradient descent. Furthermore, we normalize the scale
of the internal coordinates with the fixed parameters \SI{0.005}{\nano\metre}
for bond lengths, $0.15$ rad for bond angles, and $0.2$ rad for dihedral angles.
The circular periodic coordinates (those with freely rotating dihedral angles)
were not scaled. This internal coordinate representation is identical to the
representation used by \citeauthor{midgleyFlowAnnealedImportance2023} \yrcite{midgleyFlowAnnealedImportance2023}.

\subsubsection{Starting Dataset}
To form the starting dataset we use a short MD trajectory of \SI{50}{\pico \second}
(\num{50000} potential energy evaluations). Every 10th frame of this trajectory is used 
in the starting dataset.

\subsubsection{Architecture}
\label{SI:sec:aldp:architecture}

\paragraph{Normalizing Flow.}

\begin{figure*}[ht]
\vskip 0.2in
\begin{center}
\centerline{\includegraphics{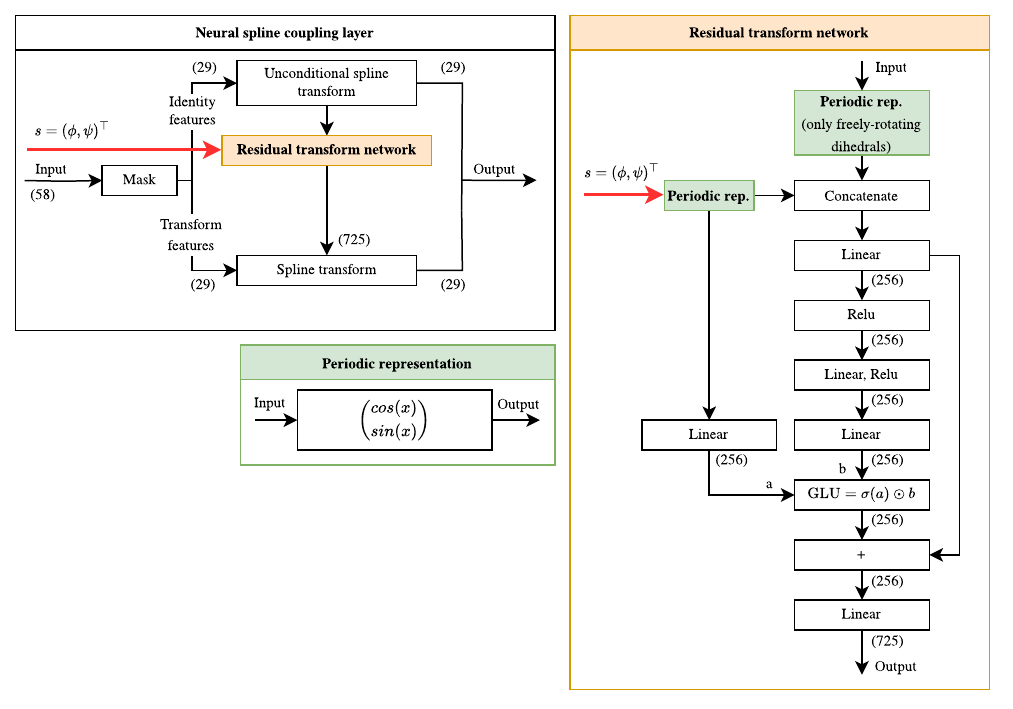}}
\caption{The architecture of the neural spline coupling layers used for alanine
dipeptide. The input of each coupling layer is split into two parts, where the
identity part is used to calculate the spline transform parameters of the
transform part using a residual network (orange). The shown architecture is very
similar to the one in \cite{midgleyFlowAnnealedImportance2023}.}
\label{fig:SI:aldp_architecture}
\end{center}
\vskip -0.2in
\end{figure*}

Our architecture is very similar to the one by
\citeauthor{midgleyFlowAnnealedImportance2023} \yrcite{midgleyFlowAnnealedImportance2023}, which builds upon the \emph{normflows}
python package \cite{stimperNormflowsPyTorchPackage2023}. We use a normalizing flow built from
12 monotonic rational-quadratic spline coupling layers
\cite{durkanNeuralSplineFlows2019}. A monotonic rational-quadratic spline
coupling layer maps the interval $[-B,B]$ to $[-B,B]$ using
monotonically increasing rational-quadratic functions in $K$ bins.

We use $K=8$ bins and treat the freely rotating dihedral angles with periodic
boundaries in the range $[-\pi,\pi]$ \cite{rezendeNormalizingFlowsTori2020a}. For
these, a uniform latent distribution in $[-\pi,\pi]$ is used. The non-periodic
coordinates use $B=5$, an identity mapping outside of the range $[-B,B]$, and a
standard Gaussian latent distribution.

We group two consecutive coupling layers to use opposite random masks to define the
identity and transform features. After such a pair of coupling layers, we apply a
random shift of $\pi \cdot U(0.5,1.5)$ with subsequent wrapping in $[-\pi,\pi]$
to the coordinates with periodic boundaries.

Each coupling layer uses a residual network as the parameter network of the
spline transformation. The detailed architecture can be found in Figure
\ref{fig:SI:aldp_architecture}. The circular
coordinates (which include the CG variables $\phi$ and $\psi$) use the
periodic representation $ (\cos(\eta), \sin(\eta))^\top $ for each periodic
variable $\eta$.

We found that training of the grid conditioning experiments is more stable and 
yields better results when not modeling the topology of $\phi$ and $\psi$ explicitly
in the flow, which is why we removed the periodic conditioning representation and 
simply condition the flow on $\phi$ and $\psi$ for the grid conditioning experiments.

\paragraph{PMF.}
To predict the PMF and its standard deviation, we use an ensemble of 10 fully
connected neural networks with layer dimensions $[4, 256, 128, 32, 1]$ and
sigmoid hidden layer activation functions. Since $\phi$ and $\psi$ are
$2\pi$-periodic, we use the input representation $(\cos \phi, \sin \phi, \cos
\psi, \sin \psi)^\top$.

\subsubsection{Hyperparameters}

\paragraph{Normalizing Flow.}
\begin{itemize}

\item Training by example: The initial training by example is performed using
the Adam optimizer with a learning rate of \num{1e-4} and a batch size of $256$
for $50$ epochs.

\item Training by energy:
\begin{itemize}
    \item As discussed in Section~\ref{sec:methods}, we found that removing a
    few of the highest loss values from each batch significantly improves training
    stability. Therefore, we removed the highest $5$ loss values.
    
    \item Active learning experiments
    \begin{itemize}
    \item Adam optimizer with batch size $64$
    \item Training for $50$ epochs in each iteration
    \item Learning rate is linearly warmed up from $0$ to \num{1e-5} in the first $15$ epochs 
    of the initial iteration (directly after training by example) and in the first $11$ epochs 
    for all subsequent AL iterations.
    \item Gradient norm clipping with value $100$ for the first $15$ epochs of
    the initial iteration (directly after training by example) and in the first
    $11$ epochs for all subsequent iterations. Then, gradient norm clipping with
    value $1000$ is used.
    \end{itemize}

    \item Grid conditioning experiments
    \begin{itemize}
    \item 80\% of the CG grid points (100x100) are used for training, 20\% for testing.
    \item Adam optimizer with batch size $64$
    \item Training for $100$ epochs in total
    \item Learning rate is linearly warmed up from $0$ to \num{1e-4} in the first $30$ epochs .
    \item Gradient norm clipping with a value of $100$ for the first $30$ epochs,
    then with a value of $1000$
    \end{itemize}
\end{itemize}

\end{itemize}

\paragraph{PMF.}
To train the PMF ensemble, we use the Adam optimizer with a learning rate of \num{5e-4}
and batch size $256$. Training is performed for $1500$ epochs. Due to possible
atom clashes and numerical instabilities \cite{koblentsPopulationMonteCarlo2015,dibakTemperatureSteerableFlows2022} 
when sampling from the flow, when evaluating Equation \ref{eq:F_main}
for each CG configuration $s$, we clip the highest $3$ of the $30$ values in the
expectation value to the lowest value among them. We find empirically that this
approach yields very accurate PMF values.

Each model in the ensemble receives a random fraction of
\SI{80}{\percent} of the current training data in the AL dataset for training
and the remaining \SI{20}{\percent} for testing.

\paragraph{Monte Carlo Sampling.}
We perform Monte Carlo Sampling in the CG space to find new high-error
configurations. We use a Gaussian proposal distribution of scale $0.1$ and an
error threshold of $0.2 \, k_\mathrm{B} T$.

In each AL iteration, we search for $30$ high-error points using $200$ MC
trajectories in parallel. We use the high-error points found in the previous AL
iteration as the starting configurations of these trajectories (the trajectories
of the first AL iteration start in the global minimum).
To obtain a more uniform coverage of the CG space, we first pick one random 
sample from these $30$ configurations and then iteratively choose the point with
the largest Euclidean distance to all already chosen points - until we have sampled a total
of $15$ points.

Each of these points is subsequently broadened with
a uniform distribution in a circle with a radius of $0.6$ rad. In this way, we sample
$200$ points for each high-error configuration, yielding $200 \cdot 15=3000$
added points in each iteration. Since each CG configuration
has $30$ copies in the dataset (see Section \ref{sec:obtaining_PMF}), this yields \num{90000} new points.
\SI{80}{\percent} are subsequently used 
for training by energy in the next iteration, \SI{20}{\percent} for calculation of a test loss.

If any of the trajectories reaches a length of \num{500000} steps, we stop the active 
learning workflow. The reported total number of MC steps do not include the steps of this final fixed-length MC
exploration that terminates the workflow due to reaching the maximum specified length.

Furthermore, if the forward KLD from the test dataset to the learned PMF in
a given iteration is smaller than \num{1e-3}, we stop the active learning workflow. When applying 
the presented methodology to systems without a ground-truth dataset in the future, one will have 
to solely rely on the maximum length or other more sophisticated stopping criteria.

\subsubsection{Training Stability}
\label{SI:sec:flow_training}

\begin{figure*}[ht]
\vskip 0.2in
\begin{center}
\centerline{\includegraphics[width=0.6\textwidth]{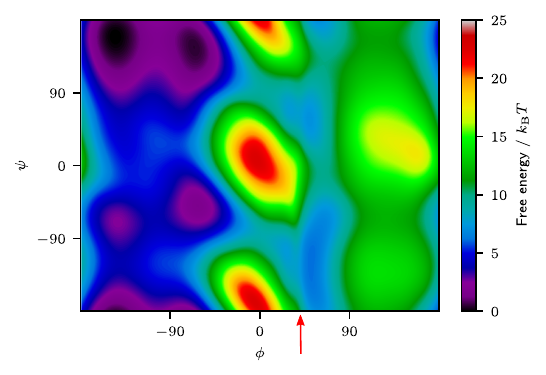}}
\caption{Small artifact (vertical line at $\phi \approx \SI{40}{\degree}$) that appears in the PMF during the active learning workflow 
applied to alanine dipeptide.}
\label{fig:SI:aldp_PMF_artifact}
\end{center}
\vskip -0.2in
\end{figure*}

As discussed in Section~\ref{sec:methods}, we found
that for alanine dipeptide removing the 5 highest loss values from each batch of 64 samples yields more 
stable experiments than energy regularization. While this yields stable AL workflows most of the time,
we observe small artifacts in approximately one out of eight AL experiments (see Figure~\ref{fig:SI:aldp_PMF_artifact}).
These artifacts are in the form of vertical lines, mostly but not always at $\phi \approx \SI{40}{\degree}$.
We found that optimizing training hyperparameters to get more stable experiments helps the problem, but does 
not completely remove it when repeating experiments many times. Therefore, we performed a total of 8 AL experiments
with different random seeds in the alanine dipeptide system to obtain reliable performance metrics in Table~\ref{table:aldp_results}.

For the grid conditioning experiments, we observed similar artifacts for some choices of hyperparameters, but the artifacts do not occur with the presented hyperparameters.

Since the number of sampled molecules of R-chirality (which are filtered) spikes at the same 
time as such an artifact typically occurs, we suspect that the artifact appears due to the hard cutoff of the chirality filtering. This is an effect of the chosen CG mapping, which has ambiguous chirality for a given $\phi$-$\psi$-configuration. If one constructs the internal coordinate representation in such a way that only structures of L-chirality can be sampled (in combination with constraining the sampling range of the respective degree of freedom in the flow \cite{rezendeNormalizingFlowsTori2020a}), the described effects should be avoidable in the future.

\subsubsection{Performance Analysis}
\label{SI:sec:aldp_performance}

\begin{figure*}[ht]
\vskip 0.2in
\begin{center}
\centerline{\includegraphics{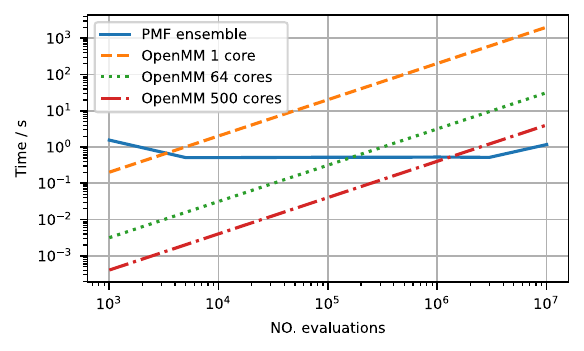}}
\caption{Benchmark that compares the time to evaluate the PMF ensemble on the
GPU and the target potential energy and force using OpenMM on CPU cores. The
OpenMM time was evaluated using the performance-optimized OpenMM CPU platform on
a single core of an Intel Xeon Platinum 8368. Based on this time, we calculated
the time for higher core numbers, assuming perfect scaling. The PMF ensemble was
evaluated on a single NVIDIA A100 40 GB GPU.}
\label{fig:SI:aldp_benchmark}
\end{center}
\vskip -0.2in
\end{figure*}

Here, we compare the performance of the MC simulations in our AL experiments
with running all-atom MD using OpenMM. For this, we compare the time over the number
of evaluations in Figure~\ref{fig:SI:aldp_benchmark}. As one can see, it takes more 
than 500 CPU cores to match the sampling speed of the PMF on the GPU - considering one 
performs enough trajectories in parallel.

As already discussed in the main text, running AL is not strictly necessary in
this low-dimensional 2D CG space, since one can simply cover it uniformly as
done in the grid conditioning experiments. Furthermore, the implicit solvent
simulation of alanine dipeptide in OpenMM is already quite fast, resulting in a
similar execution time when evaluating the PMF only once. While the nature of
parallelization of the GPU still allows faster sampling of the PMF compared to
MD in practice, one can expect an even larger difference when going to all-atom
systems that take longer to evaluate. 

In the extreme case of using a more expensive method such as density functional
theory (DFT) calculations on the all-atom side, sampling the CG surrogate will
be many orders of magnitude faster. One can, of course, also replace the DFT
with an all-atom ML surrogate and apply active learning also on the all-atom
side. However, also in this case, the CG PMF will be lower-dimensional, 
require less memory, and have shorter execution times. Furthermore, the CG PMF is
typically smoother than the all-atom energy surface, making larger time steps
possible \cite{jinBottomupCoarseGrainingPrinciples2022} when running, for
example, Langevin dynamics.

\begin{figure*}[ht]
\vskip 0.2in
\begin{center}
\centerline{\includegraphics[width=\textwidth]{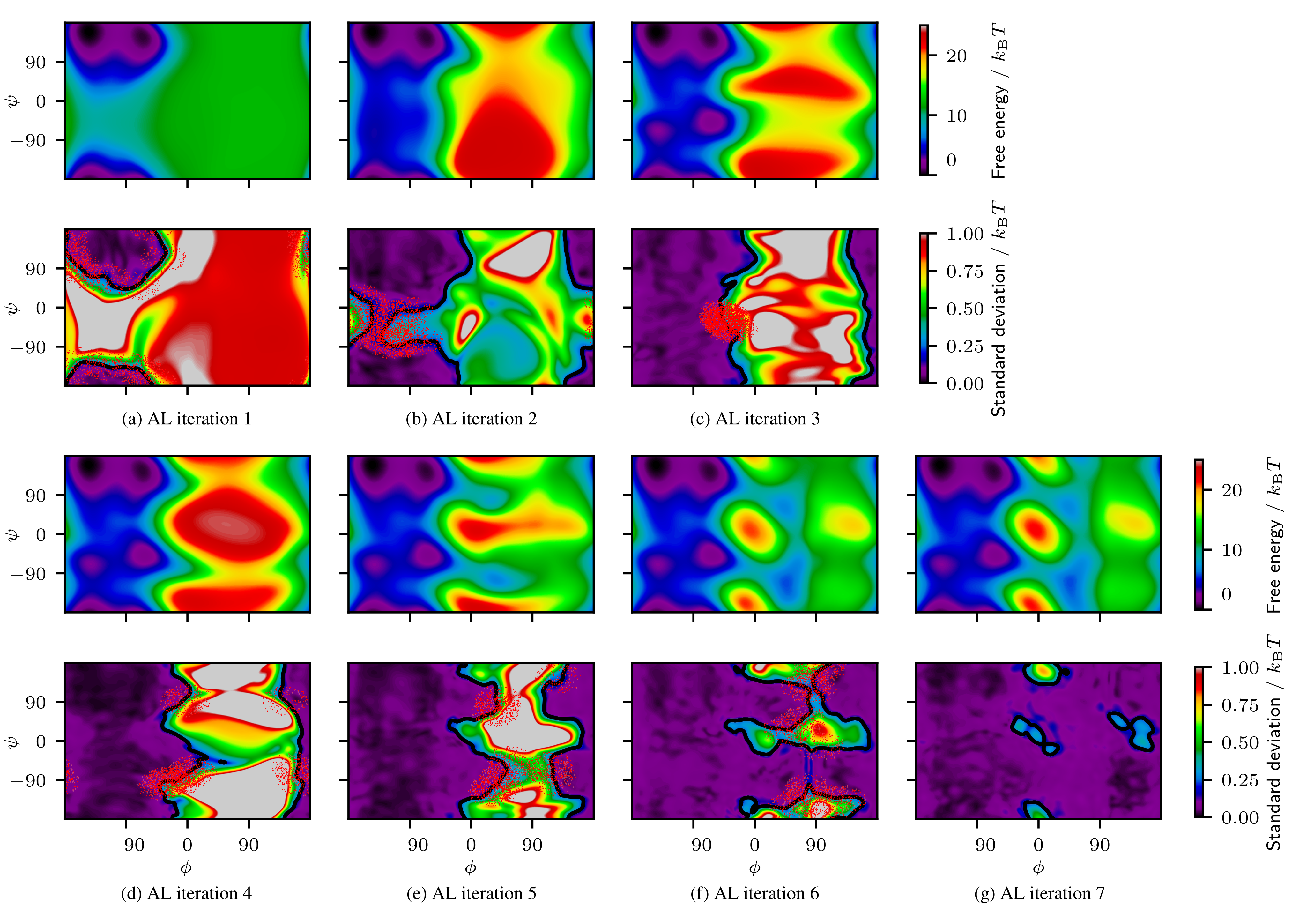}}
\caption{(a-g) AL steps for the alanine dipeptide system. Top: PMF of $\phi$ and
$\psi$ at the end of the iteration. Bottom: Standard deviation of the PMF. The
contour line of the threshold in the standard deviation when sampling new points
($0.2 \, k_\mathrm{B} T$) is drawn using a black line. Newly added points after sampling using
the PMF of this iteration are shown as red dots.}
\label{fig:SI:aldp_steps}
\end{center}
\vskip -0.2in
\end{figure*}

\subsection{Hardware Resources}
All experiments have been performed on a NVIDIA A100 40 GB GPU. For the
experiments on alanine dipeptide, we calculated the energy and gradients of the
ground-truth OpenMM system using 18 workers in parallel.

Training times of the different experiments were approximately the following:
\begin{itemize}
    \item Müller-Brown AL: \SI{35}{\minute} 
    \item Alanine dipeptide AL: \SI{36}{\hour}
    \item Alanine dipeptide grid conditioning: \SI{18}{\hour}
\end{itemize}

\end{document}